\documentclass[preprint,12pt]{elsarticle}

\usepackage{fontspec}
\usepackage{polyglossia}
\usepackage{graphicx}
\usepackage{booktabs}
\usepackage{tabularx}
\usepackage{multirow}
\usepackage{array}
\usepackage{amsmath}
\usepackage{hyperref}
\usepackage{xurl}
\usepackage[table]{xcolor}
\usepackage{csquotes}
\usepackage{placeins}
\usepackage{rotating}
\usepackage{pifont}
\usepackage[most]{tcolorbox}

\newcommand{\cmark}{\ding{51}}
\newcommand{\xmark}{\ding{55}}

\IfFontExistsTF{LibertinusSerif-Regular.otf}
  {\setmainfont{LibertinusSerif-Regular.otf}[
    BoldFont=LibertinusSerif-Bold.otf,
    ItalicFont=LibertinusSerif-Italic.otf,
    BoldItalicFont=LibertinusSerif-BoldItalic.otf]}
  {\setmainfont{texgyretermes-regular.otf}[
    BoldFont=texgyretermes-bold.otf,
    ItalicFont=texgyretermes-italic.otf,
    BoldItalicFont=texgyretermes-bolditalic.otf]}

\IfFontExistsTF{Amiri-Regular.ttf}
  {\newfontfamily\pashtofont[Script=Arabic,Scale=1.1]{Amiri-Regular.ttf}[
    BoldFont=Amiri-Bold.ttf]}
  {\newfontfamily\pashtofont[Script=Arabic,Scale=1.1]{FreeSerif.otf}[
    BoldFont=FreeSerifBold.otf]}

\usepackage{newunicodechar}
\newunicodechar{ټ}{{\pashtofont ټ}}
\newunicodechar{ډ}{{\pashtofont ډ}}
\newunicodechar{ڼ}{{\pashtofont ڼ}}
\newunicodechar{ړ}{{\pashtofont ړ}}
\newunicodechar{ښ}{{\pashtofont ښ}}
\newunicodechar{ږ}{{\pashtofont ږ}}
\newunicodechar{ځ}{{\pashtofont ځ}}
\newunicodechar{څ}{{\pashtofont څ}}
\newunicodechar{ۍ}{{\pashtofont ۍ}}
\newunicodechar{ئ}{{\pashtofont ئ}}
\newunicodechar{ې}{{\pashtofont ې}}
\newunicodechar{ۀ}{{\pashtofont ۀ}}
\newunicodechar{ګ}{{\pashtofont ګ}}
\newunicodechar{ژ}{{\pashtofont ژ}}
\newunicodechar{ہ}{{\pashtofont ہ}}
\newunicodechar{ا}{{\pashtofont ا}}
\newunicodechar{ب}{{\pashtofont ب}}
\newunicodechar{پ}{{\pashtofont پ}}
\newunicodechar{ت}{{\pashtofont ت}}
\newunicodechar{ج}{{\pashtofont ج}}
\newunicodechar{ح}{{\pashtofont ح}}
\newunicodechar{خ}{{\pashtofont خ}}
\newunicodechar{د}{{\pashtofont د}}
\newunicodechar{ر}{{\pashtofont ر}}
\newunicodechar{ز}{{\pashtofont ز}}
\newunicodechar{س}{{\pashtofont س}}
\newunicodechar{ش}{{\pashtofont ش}}
\newunicodechar{ص}{{\pashtofont ص}}
\newunicodechar{ط}{{\pashtofont ط}}
\newunicodechar{ع}{{\pashtofont ع}}
\newunicodechar{غ}{{\pashtofont غ}}
\newunicodechar{ف}{{\pashtofont ف}}
\newunicodechar{ق}{{\pashtofont ق}}
\newunicodechar{ک}{{\pashtofont ک}}
\newunicodechar{ل}{{\pashtofont ل}}
\newunicodechar{م}{{\pashtofont م}}
\newunicodechar{ن}{{\pashtofont ن}}
\newunicodechar{و}{{\pashtofont و}}
\newunicodechar{ه}{{\pashtofont ه}}
\newunicodechar{ی}{{\pashtofont ی}}
\newunicodechar{ي}{{\pashtofont ي}}
\newunicodechar{آ}{{\pashtofont آ}}

\setdefaultlanguage{english}
\newcommand{\ps}[1]{{\pashtofont #1}}

\begin{document}

\begin{frontmatter}

\title{PashtoTTS-Bench: automated screening for low-resource\\
non-Latin-script text-to-speech}

\author[ir]{Hanif Rahman}
\ead{hanif@hanifrahman.com}
\address[ir]{Independent Researcher}

\begin{abstract}
Text-to-speech (TTS) evaluation for low-resource non-Latin-script languages can
fail when it relies on a single ASR round-trip word error rate (WER). A system
may produce no audio, speak a neighbouring language, preserve target-script text
only in an ASR transcript, or sound unnatural to native listeners. We introduce
INSV (Intelligibility, Naturalness, Script fidelity, and Verification), a
reporting framework that separates these cases. This paper reports INSV-A, the
automated screening subset: synthesis completion, ASR WER/CER, transcript
Script Fidelity Rate, and audio language identification. Native MOS and
phonetic annotation are specified but not claimed in this release.

We instantiate INSV-A as PashtoTTS-Bench, a dated benchmark for Pashto TTS. The
April--May 2026 run evaluates Edge GulNawaz, Edge Latifa, OmniVoice clone,
OmniVoice auto, and an Urdu negative control on 200 FLEURS and 200 filtered
Common Voice 24 prompts. Under the independent
\texttt{omniASR\_CTC\_300M\_v2}, OmniVoice auto has the lowest WER (24.1\%
FLEURS, 27.4\% CV24), followed by Edge GulNawaz (32.8\%, 39.5\%), Edge Latifa
(35.6\%, 47.7\%), and OmniVoice clone (45.4\%, 34.8\%). WER below the
natural-speech baseline reflects clean synthetic audio and should not be read as
better than native speech. Whisper Large V3 returns 0.0\% Pashto labels on
checked Pashto TTS audio, while MMS-LID-4017 and SpeechBrain VoxLingua107
separate Pashto outputs from the Urdu control. The release provides provider
metadata, per-sentence scores, LID audits, failure logs, and scripts for adding
systems.

\end{abstract}


\end{frontmatter}

\section{Introduction}
\label{sec:intro}

\subsection{Why INSV is needed}

Text-to-speech evaluation for low-resource languages often reduces quality to a
single intelligibility number. A common protocol synthesises text to audio,
transcribes the audio with ASR, and reports word error rate (WER) between the
input text and the ASR hypothesis \citep{Taylor2021}. This protocol is
reproducible and cheap, but it misses failures that matter for non-Latin-script
languages.

A multilingual TTS system may accept Pashto input and produce Urdu or Dari
speech. It may preserve common Perso-Arabic letters but collapse Pashto-specific
letters such as ټ, ډ, ڼ, ړ, ښ, ږ, ځ, and څ. It may pronounce the right words with
flat prosody that native speakers judge as unnatural. In all three cases, a WER
score alone either hides the failure type or makes it hard to interpret.

We propose INSV as a four-part reporting framework for low-resource
non-Latin-script TTS. INSV separates intelligibility, naturalness, script
fidelity, and language verification. These dimensions answer four separate
questions: whether the output can be transcribed as the target text, whether it
sounds natural to native listeners, whether the recogniser transcript stays in
the target script, and whether the output is actually in the target language.

The present paper reports INSV-A, the automated screening part of INSV. INSV-A
does not replace native listening tests. It identifies systems that are ready for
full MOS and phonetic annotation, and it marks outputs whose WER should be read
with caution because the language or transcript evidence is ambiguous.

\subsection{PashtoTTS-Bench as a deep case study}

Pashto is a strong stress test for this framework. It has 60--80 million
speakers across Afghanistan, Pakistan, and diaspora communities, but it has no
published multi-system TTS benchmark. Commercial support is narrow: Microsoft
provides Pashto neural voices, while many multilingual providers expose no
explicit Pashto locale. Open-source systems often claim broad multilingual
coverage, but Pashto support is either absent, indirect, or hard to verify.

Pashto also has linguistic properties that make single-metric evaluation weak.
The language uses a Perso-Arabic script with letters that are absent from Arabic,
Persian, and Urdu. Its lateral fricatives, ښ /ɬ/ and ږ /ɮ/, are rare in global
speech resources. Its retroflex series and visually similar vowel letters create
grapheme-to-phoneme ambiguity. These features let us test whether multilingual
TTS systems have learned Pashto or have learned a nearby script and language
space.

We instantiate INSV-A as PashtoTTS-Bench. The benchmark uses 200 FLEURS Pashto
prompts and 200 filtered Common Voice 24 Pashto prompts. It also defines a
50-sentence MOS subset selected for Pashto-specific grapheme and phoneme
coverage, but this release reports automated screening metrics only. The release
is designed for repeated evaluation as commercial APIs and open-source models
change over time.

\subsection{Contributions}

This paper makes five contributions:
\begin{enumerate}
\item INSV, a reusable reporting framework for low-resource non-Latin-script
TTS, covering intelligibility, naturalness, transcript-script fidelity, and
language verification.
\item INSV-A, an automated screening protocol that reports synthesis completion,
ASR round-trip WER/CER, transcript SFR, and multi-model LID without claiming human
naturalness or native phonetic judgements.
\item PashtoTTS-Bench, the first open dated Pashto TTS screening benchmark, with
frozen prompts, provider metadata, per-sentence scores, language-ID audits,
failure logs, Fish Speech S2-Pro smoke-test provenance, and scripts for adding
new systems.
\item A five-part failure taxonomy for non-Latin-script TTS: pre-synthesis
rejection, language substitution, phoneme collapse, prosodic disfluency, and
grapheme ambiguity.
\item A reproducibility protocol that separates public scores and metadata from
generated commercial audio, so future studies can compare systems without
assuming that commercial audio redistribution is allowed.
\end{enumerate}

\subsection{Paper organisation}

\S\ref{sec:background} reviews Pashto phonology, TTS evaluation methods, prior
multilingual and low-resource TTS benchmarks, and the commercial/open-source
provider set. \S\ref{sec:framework} defines INSV and the automated INSV-A
instantiation. \S\ref{sec:setup} describes the systems, test sets, MOS design,
and statistical protocol.
\S\ref{sec:taxonomy} defines the failure taxonomy. \S\ref{sec:results} reports
the benchmark results and failure classifications. \S\ref{sec:discussion}
discusses provider coverage, script fidelity, evaluation circularity, and
dialect limits. \S\ref{sec:benchmark} documents the release package.
\S\ref{sec:conclusion} states the benchmark targets for future Pashto TTS
systems.

\subsection{Relationship to prior work}

This paper builds on two published companion studies. \citet{Rahman2026Benchmark}
establishes the Pashto ASR reference model (\texttt{pashto-asr-v3}) and its
natural-speech WER baselines, used here as the language-specific intelligibility
reference. \citet{Rahman2026SFR} defines Script Fidelity Rate
(\S\ref{sec:framework-sfr}), used to detect script collapse in the TTS pipeline.
The motivation for a dedicated Pashto TTS benchmark is the absence of any
published multi-system Pashto TTS evaluation in the literature
\citep{Besacier2014lowresource}; the present paper fills that gap with a
reproducible, frozen-prompt benchmark.

\FloatBarrier

\section{Background}
\label{sec:background}

\subsection{Pashto phonology and orthography}
\label{sec:pashto-background}

Pashto, spoken by 60--80 million people across Afghanistan and Pakistan, is written in a Perso-Arabic abjad script. The phoneme inventory comprises approximately 44 segments, including a series of retroflex consonants (ټ /ʈ/, ډ /ɖ/, ڼ /ɳ/, ړ /ɽ/) and a typologically rare pair of lateral fricatives (ښ /ɬ/, ږ /ɮ/). Eight Pashto graphemes are unique to the language and do not appear in Arabic, Persian, or Urdu: ټ, ډ, ڼ, ړ, ښ, ږ, ځ (/dz/), and څ (/ts/). This orthographic distinctiveness is a hard test for multilingual TTS systems. If the system's character-to-phoneme (G2P) model is trained on Arabic or Urdu data, these Pashto-unique codepoints can be misread or dropped.

Pashto orthography also has graphemic ambiguity at the character level: the vowel markers ي (Arabic yeh, U+064A), ې (Pashto e, U+06D0), and ی (Farsi yeh, U+06CC) are visually similar but phonologically distinct, and the letter و can be the consonant /w/ or the long vowel /uː/. Without a published Pashto pronunciation lexicon, multilingual TTS systems have no ground truth for these characters. This can produce systematic prosody and phonological errors.

A final complication is the absence of any published Pashto grapheme-to-phoneme rule set. Multilingual TTS systems must rely on cross-lingual transfer from related languages (Arabic, Urdu, Persian), which misses Pashto-specific phonotactics and the language-unique letters. For phonological reference, see \citet{Tegey1996grammar} and \citet{Penzl1954Orthography}.

\subsection{TTS evaluation methodology}
\label{sec:eval-methods}

\subsubsection{Automated evaluation}

The predominant method in the literature is \emph{ASR round-trip evaluation}:
synthesise a text sentence to audio via the TTS system, transcribe the audio
using an ASR model, and measure word error rate (WER) between the original text
and the ASR output \citep{Taylor2021}. This method is reproducible,
language-agnostic, and scalable, and it is widely used for TTS intelligibility
evaluation. It has three limitations. First, if the ASR model is trained on data
from the same TTS provider, the metric conflates TTS quality with ASR behaviour.
Second, WER is blind to a subset of severe failures: language substitution, script
or orthographic failures, and naturalness failures. Third, WER depends on the
quality of the reference ASR model, which for low-resource languages is often
unstable.

Complementary to WER, several automated, reference-free methods exist:
\begin{itemize}
\item \emph{Language identification (LID)}: Multilingual LID models such as Whisper Large V3 \citep{Radford2023}, Meta MMS-LID-4017 \citep{Pratap2024}, and SpeechBrain VoxLingua107 \citep{Ravanelli2021SpeechBrain,Valk2021VoxLingua107} can detect whether the synthesised audio is recognised as the target language. This screens for language substitution (F2 failures).
\item \emph{Script fidelity (SFR)}: Introduced by \citet{Rahman2026SFR}, the SFR metric measures the fraction of hypothesis characters that remain in the expected target script. SFR is a proxy for orthographic/script fidelity that does not require a reference transcription.
\item \emph{MOS prediction}: Automated models such as UTMOS \citep{Saeki2023} can estimate human naturalness (MOS) scores without human raters. However, these predictors are calibrated on English and Japanese data and their reliability for Pashto is unknown.
\end{itemize}

\subsubsection{Human evaluation}

The gold standard for TTS evaluation is human listening via mean opinion score
(MOS) on a 5-point scale. Listeners rate naturalness, and the mean and 95\%
confidence interval are reported under ITU-T subjective listening guidance
\citep{ITUP800,ITUP808,Streijl2016}. MOS is the most reliable measure of speech
quality, but it is labour-intensive and requires native speakers of the target
language. Inter-rater reliability should be reported via Krippendorff's alpha,
intraclass correlation (ICC), or an equivalent ordinal reliability statistic.

\subsection{Prior multilingual and low-resource TTS benchmarks}
\label{sec:prior-benchmarks}

Comparable TTS benchmarks are scarce for Pashto-sized languages. The ZeroSpeech
challenges \citep{Dunbar2017,Dunbar2021} target
unsupervised representation learning rather than TTS quality, and their
evaluation tasks do not include named language support or WER-based quality
screening. The Blizzard Challenge \citep{King2014Blizzard} and the VoiceMOS
Challenge \citep{Huang2022VoiceMOS} focus on English and Japanese, with MOS
as the primary metric; the latter specifically benchmarks MOS prediction models
rather than TTS systems. MSWC \citep{Mazumder2021MSWC} and FLEURS
\citep{Conneau2023} are spoken-word and sentence corpora for ASR, not TTS.

For languages outside the top 20 by resource level, dated TTS benchmarks with
per-system, per-sentence evaluation records do not appear in the literature.
No published multi-system Pashto TTS evaluation exists in the literature.
Across under-resourced languages generally, evaluation resources lag behind
resource creation \citep{Besacier2014lowresource}. The present paper fills that
gap with a reproducible, frozen-prompt benchmark that can be re-run as systems
change over time.

\subsection{Low-resource TTS: state of the field}
\label{sec:lr-tts}

The gap between high-resource (English, Mandarin Chinese) and low-resource language TTS is substantial. Commercially, only a handful of multilingual TTS providers offer coverage beyond the top 20--30 languages. For Pashto specifically:

\subsubsection{Commercial systems}

Microsoft Azure Neural TTS offers Pashto (ps-AF locale) with two voices:
GulNawaz (male) and Latifa (female). These are the only commercial Pashto TTS
voices verified in this study and are therefore the dated commercial baseline.
Google Cloud TTS, OpenAI TTS, and ElevenLabs are treated as candidates unless a
dated run verifies explicit Pashto support.

\subsubsection{Open-source multilingual systems}

Meta MMS-TTS \citep{Pratap2024} covers 1,100+ languages but \emph{explicitly
excludes} Pashto \citep{Rahman2026Benchmark}. Coqui XTTS
v2 supports multilingual synthesis and claims 20+ languages, but Pashto is not
among them. Emerging systems such as Fish Speech and OmniVoice require dated
verification before their Pashto behaviour can be cited as evidence.

\subsubsection{Structural gaps}

Across both commercial and open-source systems, the absence of a published Pashto
pronunciation lexicon is a blocker. Multilingual systems must either (a) rely on
cross-lingual G2P transfer from Arabic, Urdu, or Persian, (b) learn from
unlabeled data, or (c) use a multilingual character embedding that may conflate
Pashto letters with visually similar codepoints. This structural gap, combined
with Pashto's orthographic distinctiveness and typologically rare phonemes,
creates a systematic vulnerability in multilingual TTS. The present paper screens
for that vulnerability but does not claim phone-level diagnosis without native
annotation.

\FloatBarrier

\section{The INSV reporting framework}
\label{sec:framework}

Single-metric TTS evaluation is incomplete. A system may achieve low word error
rate (WER) while failing at language identification, transcript-script
preservation, or naturalness (the degree to which speech sounds native rather
than synthetic). We therefore define \textbf{INSV}, a four-part reporting
framework: \emph{I}ntelligibility (ASR round-trip accuracy), \emph{N}aturalness
(native-listener MOS), \emph{S}cript fidelity (target-script preservation in
the ASR transcript), and language \emph{V}erification (LID confirmation that the
output is in the target language).

The dimensions have a dependency structure. S and V should be checked before WER
is interpreted: if ASR transcribes the audio in a different script or if LID
points away from the target language, round-trip edit distance may be measuring a
category error rather than intelligibility. N requires native listeners and cannot
be replaced by an automatic score.

This paper instantiates the automated part of INSV as \textbf{INSV-A}. INSV-A
reports S, V, and I plus synthesis completion. It does not claim to measure N,
and it treats phone-level failure labels as hypotheses unless native phonetic
annotation is available.

\begin{table}[t]
\centering
\caption{INSV reporting dimensions for low-resource non-Latin-script TTS. This
paper reports the automated INSV-A subset (S, V, I and synthesis completion);
naturalness requires native-speaker MOS.}
\label{tab:insv-overview}
\small
\begin{tabularx}{\columnwidth}{p{0.08\columnwidth}p{0.21\columnwidth}p{0.26\columnwidth}X}
\toprule
\textbf{Dim.} & \textbf{Name} & \textbf{Metric} & \textbf{Method in this release} \\
\midrule
S & Script fidelity    & Transcript SFR (0--1) & Unicode block analysis over ASR output \\
V & Verification       & LID rate (\%)        & MMS-LID-4017, SpeechBrain VoxLingua107, and native labels where available. Whisper LV3 is retained as a diagnostic only$^\dagger$ \\
I & Intelligibility    & WER, CER, Perfect\% & ASR round-trip with backend labels and confidence intervals \\
N & Naturalness        & MOS (1--5 scale)     & Specified protocol; no MOS result in INSV-A \\
\bottomrule
\end{tabularx}
\smallskip\raggedright\footnotesize
$\dagger$Whisper LV3 assigns 0.0\% Pashto labels on native-validated Pashto audio; the
Pashto token exists in its vocabulary but receives negligible probability due to
near-zero Pashto training data (\S\ref{sec:discussion-lid}). Whisper is therefore
excluded from Pashto-ID rate calculations.
\end{table}

\subsection{Intelligibility: ASR round-trip evaluation}
\label{sec:framework-intelligibility}

The intelligibility dimension measures whether the synthesised speech is phonologically correct and intelligible to ASR systems trained on the target language. Given a source text $T$, a TTS system $S$ synthesises audio $A = S(T)$. An ASR model $R$ transcribes $A$ to a hypothesis $H = R(A)$. Word error rate (WER) is computed as:
$$\text{WER} = \frac{\text{edit\_distance}(T_{\text{norm}}, H_{\text{norm}})}{|T_{\text{norm}}|}$$
where $T_{\text{norm}}$ and $H_{\text{norm}}$ are normalised forms (NFC, diacritics removed, punctuation stripped). Character error rate (CER) is the character-level equivalent. We also report the percentage of utterances with zero WER (Perfect\%) and those with WER $\leq 10\%$ (Low-error\%), as these are more interpretable for systems targeting real-world use.

To reduce dependence on a single recogniser, the full benchmark protocol uses
three practices:
\begin{enumerate}
\item \emph{Validated ASR reference model}: We use a language-specific ASR model
(\texttt{pashto-asr-v3}; \citealt{PashtoW2VBERT2024}), vetted on a held-out
test set, rather than a generic multilingual model.
\item \emph{Independent ASR checks}: The protocol records every recogniser under
its backend label. Whisper Large V3 \citep{Radford2023} is retained as a stress
baseline because \citet{Rahman2026Benchmark} documents Whisper's Pashto LID
collapse on Pashto speech. Omnilingual ASR CTC
\citep{OmnilingualASR2025} is supported as an open external recogniser with
broad low-resource coverage. Language-conditioned Omnilingual LLM variants can
be run, but they must be reported separately from unconditioned CTC runs.
\item \emph{Natural speech baseline}: We include a natural speech baseline computed
within this study, which is the WER obtained when natural Pashto speech is
transcribed by the reference ASR model. This is a reference point for
interpreting TTS output.
\end{enumerate}

All WER computations use bootstrapped 95\% confidence intervals (1,000 utterance-level resamples) to quantify uncertainty.

\subsection{Naturalness: mean opinion score}
\label{sec:framework-naturalness}

Naturalness measures the perceptual quality of the synthesised speech as if
produced by a native speaker, independent of content errors. We follow ITU-T
P.800 and P.808 for the listening survey design \citep{ITUP800,ITUP808}.
Listeners rate on a 5-point scale: 1 = Bad, 2 = Poor, 3 = Fair, 4 = Good,
5 = Excellent.

\subsubsection{Rating protocol}

\emph{Dimension}: Naturalness (not intelligibility, which is captured by Dimension I).

\emph{Listener instructions}: ``Rate how natural the speech sounds, as if from a native target-language speaker, regardless of any content errors or mispronunciations.''

\emph{Listeners}: Native speakers of the target language. The Pashto protocol
targets 16 listeners across four counterbalanced forms. Twelve listeners is the
minimum exploratory pilot size; any MOS table below 16 raters must state that
the results are preliminary.

\emph{Stimuli}: A stratified 50-sentence subset of the benchmark set, selected
for language-specific phonetic coverage (e.g., retroflex consonants, lateral
fricatives for Pashto; see \S\ref{sec:setup-mos} for stratification criteria).

\emph{Blinding}: Stimuli are presented with anonymised filenames such as
\path{mos_*.mp3}. Each form has randomised order, repeated items for
within-rater consistency, and language-control clips that are excluded from MOS
means.

\emph{Inter-rater reliability}: Reported via Krippendorff's alpha (ordinal
weighting), with target $\alpha > 0.6$. If $\alpha < 0.5$, MOS results should
be reported as unreliable.

\emph{Automated proxy}: UTMOS is a public automatic MOS predictor
\citep{Saeki2023}. It is trained primarily on English and Japanese listening-test
data. We do not include it in INSV-A because it has not been validated for
Pashto; applying an English-calibrated MOS predictor to Pashto speech would
produce scores of unknown reliability. Automatic MOS remains a future direction
once per-language calibration studies are available.

\subsection{Script fidelity}
\label{sec:framework-sfr}

Script fidelity (SFR) measures whether a transcript of synthesised speech remains
in the expected writing system. For Pashto, the expected script is Perso-Arabic:
Unicode U+0600--U+06FF plus Arabic presentation forms. A Latin transcript, an
English transcript, or an empty transcript has low SFR even if a WER score can
still be computed.

For a normalised ASR hypothesis $H$, SFR is defined as:
$$\text{SFR}(H) = \frac{|\{c \in H : c \in \mathcal{S}_{\ell}\}|}{|\{c \in H : c \text{ is countable}\}|}$$
where $\mathcal{S}_{\ell}$ is the allowed script set for language $\ell$.
A character $c$ enters the denominator if and only if it is a printable
non-whitespace codepoint that is not a combining diacritic (Unicode combining
classes $> 0$), not ASCII punctuation, and not a control character; whitespace,
standard punctuation, combining diacritics (U+064B--U+065F, U+0670), and kashida
(U+0640) are excluded from both numerator and denominator. For Pashto,
$\mathcal{S}_{\ell}$ covers Unicode blocks U+0600--U+06FF (Arabic and extensions)
and presentation forms U+FB50--U+FDFF and U+FE70--U+FEFF. The metric was
introduced by \citet{Rahman2026SFR}; we adopt it as the S dimension of INSV-A.

\emph{SFR interpretation}:
\begin{itemize}
\item $\text{SFR} = 1.0$: output is entirely in the expected script.
\item $\text{SFR} \approx 0.5$: mixed script or partial script collapse.
\item $\text{SFR} \approx 0.0$: output is in a different script (full transliteration, language substitution).
\end{itemize}

In this release, SFR is a \emph{transcript-script audit}, not proof that the TTS
acoustics preserved Pashto-specific phonemes. A high SFR means the recogniser
wrote its hypothesis in the expected script. It cannot by itself rule out
phoneme collapse, dialect mismatch, or ASR bias. For this reason, the benchmark
stores SFR per ASR backend and preserves empty transcripts explicitly.

\subsection{Language verification}
\label{sec:framework-lid}

Language verification asks whether synthesised audio is recognised as the target
language or as a neighbouring language. This release supports three automatic
LID signals and native adjudication:
\begin{enumerate}
\item \emph{Whisper Large V3}: language reporting from the ASR pipeline; Pashto-ID rate is the percentage of utterances assigned language label 'ps'.
\item \emph{MMS-LID-4017}: Meta's multilingual LID covering 4,017 languages \citep{Pratap2024}; provides a second LID signal.
\item \emph{SpeechBrain VoxLingua107 ECAPA}: a non-Meta LID model trained on
VoxLingua107 \citep{Ravanelli2021SpeechBrain,Valk2021VoxLingua107}; it provides
an architecture and data source outside the Whisper/MMS family.
\item \emph{Native LID labels}: blinded native-speaker labels are treated as the
adjudication layer for disputed automatic labels.
\end{enumerate}

No automatic LID model is treated as an oracle. If models disagree, INSV-A marks
the language status as unresolved and requires native adjudication for a final
F2 language-substitution label. This is especially important for Pashto because
Whisper can assign non-Pashto labels to native-validated Pashto speech.

\emph{Interpretation}:
\begin{itemize}
\item Pashto-ID rate $> 90\%$ with agreement across models: language likely correct; proceed to I and N analysis.
\item Pashto-ID rate $< 50\%$: likely language substitution (F2 failure); WER and MOS are not interpretable.
\item Model disagreement or Pashto-ID rate 50--90\%: unresolved; report WER with caveat and collect native labels.
\end{itemize}

\subsection{Applicability to other low-resource languages}
\label{sec:framework-transfer}

INSV-A is validated empirically on Pashto only. Structurally, adapting it to
another non-Latin-script language requires six language-specific inputs;
everything else in the pipeline transfers without modification.
Table~\ref{tab:insv-transfer-checklist} lists those inputs. Examples of script
ranges: Urdu/Pashto share U+0600--U+06FF; Ethiopic is U+1200--U+137F; Thai is
U+0E00--U+0E7F. For languages where Whisper does not include a language code,
MMS-LID-4017 or SpeechBrain VoxLingua107 serve as the LID layer.

\begin{table}[t]
\centering
\caption{Language-specific inputs required to apply INSV-A to a new language.}
\label{tab:insv-transfer-checklist}
\small
\begin{tabularx}{\columnwidth}{lX}
\toprule
\textbf{Input} & \textbf{Purpose} \\
\midrule
Frozen text set & Shared prompts; enables cross-run comparisons. \\
Target ASR model & Round-trip WER/CER for Dimension I. \\
Language-ID model(s) & Target-language verification for Dimension V. \\
Native raters & MOS ratings and error annotation for Dimensions N and F3/F5. \\
Script Unicode ranges & SFR computation for Dimension S. \\
Grapheme class list & Class-level WER screening for F3 candidates. \\
\bottomrule
\end{tabularx}
\end{table}

\FloatBarrier

\section{Experimental setup}
\label{sec:setup}

\subsection{Systems evaluated}
\label{sec:systems}

We evaluate systems that were publicly accessible through an API, command-line
package, or open model repository at the time of the run. Systems with explicit
Pashto support are evaluated as Pashto TTS systems. Systems without explicit
Pashto support are listed as candidates only when their model identifier or
inference path could not be verified for this dated run.

\begin{table*}[t]
\centering
\caption{TTS systems and inclusion status for the April--May 2026 automated run.}
\label{tab:systems}
\small
\begin{tabularx}{\textwidth}{p{0.20\textwidth}p{0.13\textwidth}p{0.18\textwidth}X}
\toprule
\textbf{System} & \textbf{Run date} & \textbf{Status} & \textbf{Access and model identifier} \\
\midrule
Edge TTS GulNawaz    & 2026-04 & included & \texttt{edge-tts} 7.2.8, \texttt{ps-AF-GulNawazNeural}; native Pashto male voice. \\
Edge TTS Latifa      & 2026-05-12 & included & \texttt{edge-tts} 7.2.8, \texttt{ps-AF-LatifaNeural}; native Pashto female voice. \\
OmniVoice clone      & 2026-05 & included & \texttt{k2-fsa/OmniVoice}; zero-shot voice cloning with Edge GulNawaz reference audio. \\
OmniVoice auto       & 2026-05-12 & included & \texttt{k2-fsa/OmniVoice}; automatic voice mode without reference audio or speaker cloning. \\
Edge Urdu Asad       & 2026-05-12 & control & \texttt{edge-tts}, \texttt{ur-PK-AsadNeural}; neighbouring-language negative control, not a Pashto TTS system. \\
Fish Speech S2-Pro   & 2026-05-15 & smoke-tested & \texttt{fishaudio/s2-pro}; full 200-sentence synthesis remains incomplete in this release. Smoke-test artifacts and setup logs are retained; quality metrics are withheld. \\
Google Cloud TTS     & --- & excluded & Pashto locale not verified during this run. \\
OpenAI TTS           & --- & excluded & No explicit Pashto locale and no provider run. \\
ElevenLabs v2/v3     & --- & excluded & No explicit Pashto locale and no provider run. \\
XTTS v2 (Coqui)      & --- & excluded & Pashto not listed among supported languages for the tested package. \\
MMS-TTS (Meta)       & --- & excluded & No released \texttt{facebook/mms-tts-ps} checkpoint. \\
\bottomrule
\end{tabularx}
\end{table*}

\subsection{Test sets}
\label{sec:testsets}

We evaluate across two test sets to assess both broadcast-quality and crowdsourced-quality speech:

\subsubsection{FLEURS Pashto (ps\_af)}

The Google FLEURS benchmark \citep{Conneau2023} provides 200 high-quality,
professionally read Pashto prompts in the local benchmark cache, spanning
broadcast news and similar domains. Sentence length ranges from 5 to 35 words.
This set is the primary test set. The natural speech WER baseline under the
language-specific reference model is 34.6\% (normalised; computed in this study;
see \S\ref{sec:results-automated}). FLEURS enables comparison with prior Pashto
ASR work \citep{Rahman2026Benchmark} and is useful for evaluating intelligibility
under controlled read-speech conditions.

\subsubsection{Common Voice 24 Pashto (filtered)}

Mozilla Common Voice 24 Pashto test data filtered to 200 prompts in the local
benchmark cache. Filtering criteria: audio duration $\geq 1.0\text{s}$, RMS
energy $\geq 0.005$, and transcription word count $\geq 3$. This set assesses
robustness to non-broadcast speech (accents, background noise, microphone
variation). The natural speech WER baseline for this set is 32.5\% (normalised;
computed in this study; see \S\ref{sec:results-automated}).

\subsubsection{Stratified MOS subset}

A 50-sentence subset stratified from FLEURS is specified for later MOS
evaluation. Stratification criteria are sentence length 5--25 words, at least
one Pashto-specific grapheme class (retroflex ټ/ډ/ڼ/ړ, lateral fricative ښ/ږ,
affricate ځ/څ, or Pashto vowel marker), and available audio for every core
Pashto TTS system. The export script creates four blinded survey forms so each
sentence appears once per rater and once per system across forms. This automated
release publishes the selection procedure and export script, but it does not
report MOS scores.

\subsubsection{Text normalisation}

All WER/CER computations use Unicode NFC normalisation, kashida removal
(U+0640), punctuation stripping, and diacritic removal (U+064B--U+065F,
U+0670). This is the standard Pashto ASR text normalisation protocol, consistent
with prior Pashto ASR evaluation \citep{Rahman2026Benchmark}.

\subsection{MOS evaluation design}
\label{sec:setup-mos}

The full protocol follows \S\ref{sec:framework-naturalness}. This section
records the Pashto-specific operational parameters.

\textbf{Stratified subset.} 50 sentences from the FLEURS prompt set, stratified
by sentence length (5--25 words) and coverage of at least one Pashto grapheme
class (retroflex, lateral fricative, affricate, Pashto vowel marker).

\textbf{Target rater count.} 16 native Pashto speakers across four
counterbalanced forms; 12 is the minimum exploratory pilot size. Rater metadata
include dialect or region and daily Pashto use. No names, addresses, or IP
addresses are collected.

\textbf{Controls.} Each listener rates two Urdu-voice clips as attention and
language checks; these are excluded from Pashto MOS means. Three repeated items
check within-rater consistency.

\textbf{Export.} Survey forms, rater instructions (in English and Pashto), and
the blinded audio export script are in \path{scripts/export_mos_survey.py} and
Supplementary Material, \S A.

\subsection{ASR models for intelligibility}
\label{sec:asr-models}

We use two ASR models. \texttt{omniASR\_CTC\_300M\_v2} is the primary
cross-system check because it is independent of all evaluated providers and was
not trained by the author. \texttt{ihanif/pashto-asr-v3} is a
language-specific reference that provides more reliable Pashto decoding but
introduces a risk that author-trained model familiarity influences the
comparison. Both models are reported; the independent model takes precedence
wherever the two disagree on ordering.

\textbf{Conflict of interest note.}\label{sec:conflict} The author trained pashto-asr-v3 and
created the Microsoft Edge Pashto TTS training dataset. No commercial
relationship exists with OmniVoice or any other evaluated provider. All
evaluation scripts and result CSVs are published so that any reader can
re-rank systems under a different recogniser.

\subsubsection{Independent cross-check: omniASR\_CTC\_300M\_v2}

\texttt{omniASR\_CTC\_300M\_v2} is an open, unconditioned CTC model with broad
low-resource coverage \citep{OmnilingualASR2025}. It is not trained by the
author and has no known relationship to the Edge TTS training data. We treat
its WER as the primary cross-system metric; it is labelled O-WER in all tables.

\subsubsection{Language-specific reference: pashto-asr-v3}

\texttt{ihanif/pashto-asr-v3} \citep{PashtoW2VBERT2024} achieves 34.6\% WER on
FLEURS natural speech (95\,\% CI 32.3--37.1\%, $n$=200; computed in this study).
The CV24 natural speech baseline (32.5\%) is taken from \citet{Rahman2026Benchmark}
because the CV24 audio dataset is not publicly redistributable; the FLEURS result
confirms model behaviour is consistent with that prior measurement. It produces more
reliable Pashto decoding than unconditioned multilingual models and is used as
a secondary diagnostic. Its WER is labelled P-WER. Pashto-asr-v3 was
fine-tuned on FLEURS training data, so FLEURS test WER may be optimistic
relative to other domains; CV24 results serve as a partial cross-domain check.

\subsubsection{Diagnostic: Whisper Large V3}

Whisper Large V3 is retained for LID stress-testing only. Its language
vocabulary includes \texttt{<|ps|>} (token ID~50340), and the generation config
leaves this token available, but Whisper returns 0.0\% Pashto labels in practice.
This reflects near-zero Pashto representation in Whisper's training corpus
\citep{Radford2023}: the token exists but the model assigns it negligible
probability, routing Pashto audio to neighbouring language tokens. Whisper WER
is retained as a diagnostic baseline but is not used as an intelligibility
metric or LID signal.

\subsection{Statistical analysis}
\label{sec:stats}

WER and CER results are accompanied by 95\% bootstrap confidence intervals
(1,000 utterance-level resamples). MOS results, once collected, must report
per-system confidence intervals, rater count, counterbalanced form assignment,
and inter-rater reliability. Pairwise MOS significance tests are exploratory
below 16 native raters.

\FloatBarrier

\section{A proposed failure typology for non-Latin-script TTS}
\label{sec:taxonomy}

TTS failures for non-Latin-script languages often fall into one of five
categories, arranged by severity and detectability. The five modes below
constitute a \emph{proposed typology} derived from manual inspection of
PashtoTTS-Bench outputs and from the linguistic properties of Pashto. They are
not a validated taxonomy in the sense of a systematic study across multiple
languages; comprehensive cross-language validation is left for future work. The
typology is designed as a reusable reference for any non-Latin-script language,
with concrete examples grounded in Pashto linguistic challenges. The five modes
map to the INSV framework: F1 and F2 can be screened with Dimensions S and V;
F3 and F5 require native phonetic or orthographic annotation; F4 requires human
naturalness evaluation.

\begin{table*}[t]
\centering
\caption{Taxonomy of TTS failure modes for non-Latin-script languages. Detection method maps to the INSV framework dimension (Table~\ref{tab:insv-overview}).}
\label{tab:taxonomy}
\small
\begin{tabularx}{\textwidth}{p{0.15\textwidth}X X}
\toprule
\textbf{Mode} & \textbf{Definition} & \textbf{INSV detection} \\
\midrule
F1: Pre-synthesis rejection &
  System refuses target-language input, returns silence, or falls back before synthesis. &
  Missing audio, API error, or undefined SFR. Severity: terminal. \\[3pt]
F2: Language substitution &
  System accepts input but speaks a different language, often Urdu, Dari, Arabic, or English. &
  Target LID fails; WER is treated as non-interpretable. Severity: severe. \\[3pt]
F3: Phoneme collapse &
  System neutralises language-specific sounds, including Pashto lateral fricatives and retroflexes. &
  Phoneme-class WER identifies candidates; native annotation confirms the pattern. Severity: moderate. \\[3pt]
F4: Prosodic disfluency &
  Output is intelligible but has flat pitch, stress errors, or unnatural rhythm. &
  MOS is low relative to WER. Severity: moderate. \\[3pt]
F5: Grapheme ambiguity &
  System conflates orthographic variants, for example ي vs.\ ې or و as vowel vs.\ consonant. &
  CER rises relative to WER; native annotation identifies the grapheme class. Severity: low--moderate. \\
\bottomrule
\end{tabularx}
\end{table*}

\subsection{F1: Pre-synthesis rejection}

Systems may refuse to synthesise Pashto input without producing any output. This occurs when the language code (ps or ps-AF) is absent from the system's supported locale list, or when input validation rejects non-Latin scripts without fallback. For example, if a system supports only a fixed locale list and Pashto is absent, the synthesis request may fail silently or return an HTTP 400 error. An F1 failure is the easiest to detect: SFR is undefined, audio duration is zero or near-silence, and API logs show an error code. F1 is language-universal: any TTS system can reject any input script.

In PashtoTTS-Bench, F1 is recorded from provider logs rather than inferred from
provider documentation.

\subsection{F2: Language substitution}

The system accepts Pashto input but produces speech in a different language. This
is a severe failure because WER will often be near 100\%, and the system appears
broken when in fact it is generating output in the wrong language. An example:
given the Pashto sentence \ps{ته څوک ی؟} (``Who are you?''), a system might
produce speech that, when transcribed by Whisper, yields \ps{تم کون ہو} in Urdu
(``Who are we?'') instead of the Pashto equivalent. The resulting WER is
meaningless because it compares two different languages.

F2 is screened via language identification (Dimension V): Whisper Large V3,
MMS-LID, or SpeechBrain VoxLingua107 may assign a non-Pashto language label
(typically Urdu or Dari for Pashto inputs to cross-lingual systems). F2 is
language-universal but manifests differently for each language: Persian TTS
systems may substitute Farsi for Persian, Arabic systems may substitute Modern
Standard Arabic for dialect variants. In this benchmark, multilingual systems
without explicit Pashto support are treated as F2 candidates until native labels
confirm or reject the failure.

\subsection{F3: Phoneme collapse}

The system produces speech in approximately the right language but neutralises Pashto-unique phonemes. It replaces them with the nearest phoneme from a larger language (e.g., Arabic or Urdu). Pashto's lateral fricatives (ښ /ɬ/, ږ /ɮ/) are typologically rare and absent from Arabic and Urdu; thus, multilingual systems trained primarily on Arabic/Urdu data may collapse these to the nearest sibilants (ش /ʃ/, ژ /ʒ/). Similarly, Pashto retroflexes (ټ /ʈ/, ډ /ɖ/, ڼ /ɳ/, ړ /ɽ/) may be neutralised to plain alveolars (ت /t/, د /d/, ن /n/, ر /ɾ/) if the system's training data covers Urdu or Hindi retroflexes but not Pashto's distinct retroflex inventory.

Phoneme collapse can be screened via phoneme-class WER: compute WER on subsets of
utterances, grouped by the graphemes they contain. For Pashto, if the lateral
fricative subset (utterances containing ښ or ږ) has WER much higher than the
overall average, F3 becomes a candidate label. The label is confirmed only when a
native annotator hears the expected phone replaced by another phone. F3 is partly
language-specific: lateral fricatives are unusual in global TTS data, while
retroflex contrasts occur in several South Asian languages but differ across
phonological systems.

\subsection{F4: Prosodic disfluency}

The system produces lexically correct speech (low WER, high Perfect\%) but the
output is unnaturally flat, with monotone or unusual pitch contours, unnatural
stress placement, or erratic rhythm. A Pashto speaker would understand the
output but would rate it as ``Poor'' or ``Fair'' on the MOS scale. Pashto is a
stress-timed language with lexical stress \citep{Tegey1996grammar}; a system
that ignores word stress and produces uniform pitch will sound robotic and
non-native. F4 is detectable only via human evaluation (Dimension N): MOS can be
low even when WER is acceptable.

F4 is language-universal but the specific prosodic defects depend on language
phonology. F4 must be judged by native listeners because automatic metrics do
not reliably measure stress and rhythm.

\subsection{F5: Grapheme ambiguity failure}

Pashto orthography contains several homoglyphic pairs: ي (Arabic yeh, U+064A, /j/) vs.\ ې (Pashto e, U+06D0, /e/) vs.\ ی (Farsi yeh, U+06CC, /i/); and و (waw, U+0648), which can be the consonant /w/ or the vowel /uː/. Systems that do not include a Pashto-specific pronunciation lexicon must rely on cross-lingual grapheme-to-phoneme rules or character embeddings. If the system confuses ي with ې, it may produce /j/ where /e/ is required. The output can preserve much of the target phonology while using the wrong orthographic convention. For example, the word ``پې'' (``then'') might be read as ``پي'' (a rare interjection or name). The semantic and prosodic interpretation changes.

F5 can be screened via character error rate (CER): if CER is high relative to
WER, the item may contain grapheme-level substitutions that do not dominate the
word-level score. Confirmation requires native annotation; cross-backend
consensus can raise suspicion but cannot substitute for a native listener
confirming the substitution is systematic rather than coincidental. In
PashtoTTS-Bench, three FLEURS items show a consistent U+06CC → U+06D0
substitution under both pashto-asr-v3 and omniASR. This pattern is consistent
with F5 and motivates native phonetic review, but it should not be reported as
a confirmed F5 label without such review (see Supplementary Material, \S C).

\subsection{Taxonomy universality}

F1, F2, and F4 are fully language-agnostic. Any non-Latin-script language can experience pre-synthesis rejection (F1), language substitution (F2), and prosodic degradation (F4). F3 (phoneme collapse) is universal in principle but manifests language-specifically: the vulnerable phoneme classes depend on typological rarity in the training data. F5 (grapheme ambiguity) is partly language-specific (depends on the target language's orthographic ambiguities) and partly universal (any language with ambiguous graphemes may experience it).

To apply this taxonomy to a new language (e.g., Urdu), one would:
\begin{enumerate}
\item Use Dimensions I, N, V, S to measure system performance across the four INSV dimensions.
\item For each system showing suboptimal performance, determine which failure modes explain the shortfall.
\item Use phoneme-class WER analysis (from Dimension I) to screen F3; use LID
(Dimension V) to screen F2; use MOS (Dimension N) to detect F4.
\item Annotation of error examples (F5 and phoneme collapse details) requires native speaker expertise specific to the target language.
\end{enumerate}

\FloatBarrier

\section{Results}
\label{sec:results}

This section reports automated INSV-A results from the April--May 2026 dated
run. It covers synthesis completion, ASR round-trip intelligibility, transcript
SFR, and language-ID screening. It does not report MOS, independent native
language adjudication, or native phonetic annotation.

\subsection{Automated benchmark results}
\label{sec:results-automated}

\paragraph{Natural speech baseline under omniASR}
We ran \texttt{omniASR\_CTC\_300M\_v2} on 200 natural-speech FLEURS Pashto test
utterances (same medium-length selection logic as the benchmark prompts) within
this study independently. The result is 47.9\,\% WER (95\,\% CI
45.6--50.1\,\%; zero perfect-WER utterances). Natural speech therefore scores
nearly twice the WER of OmniVoice auto (24.1\,\%) and substantially above
Edge GulNawaz (32.8\,\%) under the same unconditioned recogniser. This confirms
that the ranking advantage of TTS systems over natural speech on O-WER is a
property of acoustic cleanliness, not phonological quality. The pashto-asr-v3
baseline (34.6\,\% FLEURS / 32.5\,\% CV24) is lower because that model is
tuned for Pashto and handles natural speech variance more robustly.

\begin{table*}[t]
\centering
\caption{PashtoTTS-Bench automated screening results, April--May 2026 dated
run. Synth.\ = non-zero audio files synthesised. ASR $n$ = usable
transcriptions (identical across both recognisers except where noted).
O-WER = \texttt{omniASR\_CTC\_300M\_v2} WER (independent, primary check;
95\% bootstrap CI in parentheses).
P-WER = \texttt{pashto-asr-v3} WER (language-specific reference).
SFR = transcript Script Fidelity Rate over the O-WER hypothesis. MMS and SB =
Pashto-ID rates from MMS-LID-4017 and SpeechBrain VoxLingua107. The Urdu row
is a negative control; Fish Speech S2-Pro is listed as a planned open-model
condition and excluded from ranked results in this release.}
\label{tab:results-main}
\scriptsize
\resizebox{\textwidth}{!}{
\begin{tabular}{llrrlrrrrl}
\toprule
\textbf{System} & \textbf{Source} & \textbf{Synth.} &
\textbf{ASR $n$} & \textbf{O-WER (95\% CI)} & \textbf{P-WER} &
\textbf{SFR} & \textbf{MMS} & \textbf{SB} &
\textbf{Status} \\
\midrule
\multirow{2}{*}{Edge GulNawaz}
  & FLEURS & 200 & 200 & 32.8 (±2.0) & 26.3 & 1.000 & 97.0 & 100.0 & native Pashto voice \\
  & CV24   & 200 & 200 & 39.5 (±3.4) & 32.7 & 1.000 & 65.0 & 98.0 & native Pashto voice \\
\midrule
\multirow{2}{*}{Edge Latifa}
  & FLEURS & 200 & 200 & 35.6 (±1.9) & 27.0 & 1.000 & 99.5 & 100.0 & native Pashto voice \\
  & CV24   & 200 & 200 & 47.7 (±3.8) & 34.4 & 1.000 & 82.0 & 96.0 & native Pashto voice \\
\midrule
\multirow{2}{*}{OmniVoice clone}
  & FLEURS & 195 & 193\textsuperscript{†} & 45.4 (±3.1) & 41.2 & 1.000 & 94.9 & 99.0 & partial F1; V unresolved \\
  & CV24   & 200 & 200 & 34.8 (±3.4) & 32.0 & 1.000 & 82.0 & 99.5 & V unresolved \\
\midrule
\multirow{2}{*}{OmniVoice auto}
  & FLEURS & 200 & 200 & 24.1 (±1.4) & 20.4 & 1.000 & 100.0 & 100.0 & non-clone open model \\
  & CV24   & 200 & 200 & 27.4 (±2.7) & 25.8 & 1.000 & 95.0 & 100.0 & non-clone open model \\
\midrule
\multirow{2}{*}{Edge Urdu Asad}
  & FLEURS & 200 & 200 & 90.1 (±1.1) & 77.8 & 1.000 & 9.0 & 3.0 & negative control \\
  & CV24   & 200 & 200 & 92.0 (±1.8) & 80.4 & 1.000 & 9.0 & 11.5 & negative control \\
\midrule
\multirow{2}{*}{Fish Speech S2-Pro}
  & FLEURS & --- & --- & --- & --- & --- & --- & --- & smoke-tested only \\
  & CV24   & --- & --- & --- & --- & --- & --- & --- & smoke-tested only \\
\midrule
\multicolumn{2}{l}{Natural speech (omniASR, this paper, $n$=200)}
  & 200 & 200 & 47.9 (±2.3) & --- & --- & --- & --- & FLEURS \\
\multicolumn{4}{l}{Natural speech (pashto-asr-v3, FLEURS computed this study; CV24$^{\ddagger}$)}
  & --- & 34.6\,/\,32.5 & --- & --- & --- & FLEURS / CV24 \\
\bottomrule
\end{tabular}
}
\smallskip
\raggedright\scriptsize
\textsuperscript{†}OmniVoice clone FLEURS missing indices: 13, 14, 63, 66, 73, 188, 200 (5 synthesis failures and 2 ASR failures); excluded from WER.\\
\textsuperscript{$\ddagger$}Natural-speech baselines are reference rows, not TTS systems. OmniASR FLEURS baseline and pashto-asr-v3 FLEURS baseline are computed in this study; CV24 pashto-asr-v3 baseline is from \citet{Rahman2026Benchmark}.
\end{table*}

\IfFileExists{figures/fig2_wer_comparison.pdf}{
\begin{figure}[t]
\centering
\includegraphics[width=\linewidth]{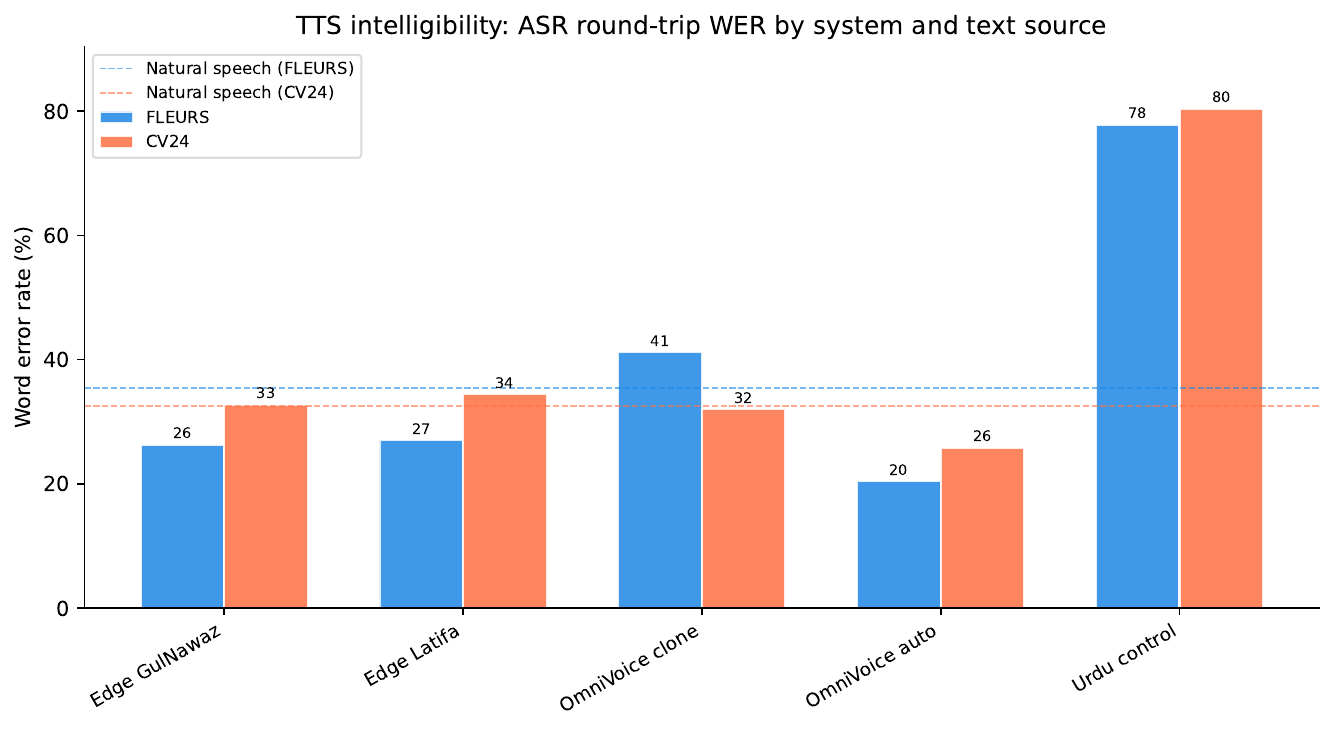}
\caption{ASR round-trip WER by system and source text set. Error bars show
95\% bootstrap confidence intervals. The dashed reference line marks the
natural-speech pashto-asr-v3 WER baseline (34.6\% on FLEURS computed this study; 32.5\% on CV24 from \citealt{Rahman2026Benchmark}).}
\label{fig:wer-comparison}
\end{figure}
}{}

\paragraph{Edge GulNawaz}
Edge GulNawaz produced 200/200 FLEURS and 200/200 CV24 files. Under the
independent omniASR model, WER is 32.8\,\% on FLEURS and 39.5\,\% on CV24
(95\,\% CI $\pm2.0$ and $\pm3.4$ respectively). The language-specific
pashto-asr-v3 gives lower absolute values (26.3\,\% / 32.7\,\%) and the same
ranking. Both transcript SFR means are 1.0. SpeechBrain Pashto-ID rates are
high (100.0\,\% / 98.0\,\%). MMS-LID is high on FLEURS (97.0\,\%) but
substantially lower on CV24 (65.0\,\%), a 33-point gap relative to SpeechBrain
that we cannot explain without native adjudication. We treat this as a
cautionary signal, leave F2 unresolved, and flag GulNawaz CV24 audio for native
LID review before the MMS result can be interpreted as a quality signal.

\paragraph{Edge Latifa}
Edge Latifa produced 200/200 files on both sets. Under omniASR, WER is
35.6\,\% on FLEURS and 47.7\,\% on CV24. Pashto-asr-v3 gives 27.0\,\% /
34.4\,\%. The omniASR CV24 gap between Latifa and GulNawaz (47.7\,\% vs.\
39.5\,\%) is wider than the pashto-asr-v3 gap (34.4\,\% vs.\ 32.7\,\%), which
suggests the independent recogniser is more sensitive to the acoustic
differences in crowdsourced-domain text. LID rates remain high.

\paragraph{OmniVoice clone and auto modes}
The cloned OmniVoice run produced 195/200 FLEURS files and 200/200 CV24 files;
7 files are missing from the FLEURS per-sentence results (5 synthesis failures,
2 further ASR failures). Missing file indices, word counts, and Pashto grapheme
class membership are documented in Table~\ref{tab:results-main} footnote.
Missing files are excluded from WER computation, not counted as errors.
Under omniASR, the clone is weak on FLEURS (45.4\,\%) but closer to Edge on
CV24 (34.8\,\%). Pashto-asr-v3 gives 41.2\,\% and 32.0\,\%.

The non-clone OmniVoice auto mode produced all 400 files. Under omniASR it
gives the lowest WER of any system: 24.1\,\% on FLEURS (95\,\% CI $\pm1.4$)
and 27.4\,\% on CV24 ($\pm2.7$). Pashto-asr-v3 gives 20.4\,\% / 25.8\,\%.
Both values are below the natural-speech pashto-asr-v3 baseline of
34.6\,\% / 32.5\,\% (FLEURS computed this study; CV24 from \citealt{Rahman2026Benchmark}).

\begin{tcolorbox}[colback=gray!8, colframe=gray!50, boxrule=0.4pt,
  left=4pt, right=4pt, top=4pt, bottom=4pt]
\textbf{Why OmniVoice auto WER is below the natural-speech baseline.}
OmniVoice auto scores 24.1\,\% O-WER on FLEURS. The same recogniser scores
natural FLEURS speech at 47.9\,\% (95\,\% CI 45.6--50.1\,\%; $n$=200;
see Table~\ref{tab:results-main}). The 23.8-point gap arises
because \texttt{omniASR\_CTC\_300M\_v2} is an unconditioned multilingual model
with no Pashto specialisation: it performs poorly on natural Pashto speech
(accent variation, spontaneous disfluencies, recording noise) but performs
substantially better on the clean, noise-free, controlled-pace synthetic audio
that OmniVoice auto produces. This is a property of the recogniser, not an
indication that OmniVoice audio is more intelligible than native speakers.
The pashto-asr-v3 model, which is tuned for Pashto, scores natural speech at
34.6\,\% and OmniVoice auto at 20.4\,\% --- a smaller gap because the tuned
model handles natural Pashto more reliably. WER rankings under either recogniser are relative screening values only;
naturalness requires native MOS. The O-WER system ordering (OmniVoice auto
$<$ GulNawaz $<$ Latifa $<$ OmniVoice clone) is treated as meaningful only
because it is reproduced under pashto-asr-v3, indicating the pattern is not
an artefact of one recogniser's behaviour.
\end{tcolorbox}

\paragraph{Fish Speech S2-Pro}
Fish Speech S2-Pro (\texttt{fishaudio/s2-pro}) is retained as a planned
open-model condition. Local attempts produced smoke-test artifacts and setup
logs, but no complete dated 200+200 synthesis run. We therefore exclude Fish
Speech S2-Pro from the ranked automated results and make no quality claim until
the full synthesis, ASR, LID, and MOS export exists.

\paragraph{Urdu negative control}
The Urdu voice produces non-empty audio for every Pashto prompt, but both
recognisers reject it strongly: 77.8--80.4\,\% WER under pashto-asr-v3 and
90.1--92.0\,\% WER under omniASR. MMS and SpeechBrain Pashto false-positive
rates are low (9.0\,\% and 3.0--11.5\,\%). This control confirms the benchmark
separates neighbouring-language speech from Pashto-targeted output.

\subsection{Language verification and script fidelity}
\label{sec:results-lid-sfr}

\paragraph{SFR was uninformative in this run.} All four evaluated systems
achieved SFR\,=\,1.0 on both test sets. This was expected: every system either
has an explicit Pashto locale (Edge voices) or uses a Pashto voice as the
cloning reference (OmniVoice). Systems with no Pashto-specific path would be
the cases where SFR adds discriminating signal --- for example, a generic
multilingual model given Pashto text without a Pashto locale. The S dimension
is retained in INSV-A because such cases are realistic in practice and SFR = 1.0
represents an empirical outcome of this run rather than a design flaw in the
metric.

\IfFileExists{figures/fig4_sfr_lid.pdf}{
\begin{figure}[t]
\centering
\includegraphics[width=\linewidth]{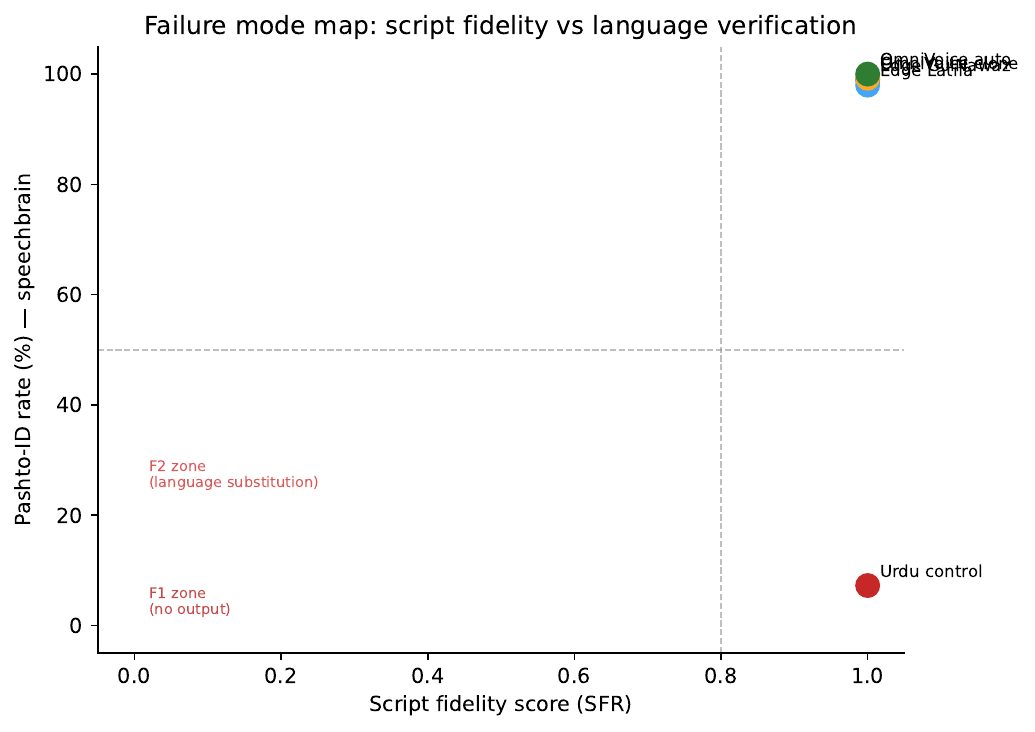}
\caption{Failure map from transcript SFR and language verification. The
vertical reference line shows a 90\,\% Pashto-ID rate, but model disagreement is
reported as unresolved rather than overridden by one model.}
\label{fig:sfr-lid-map}
\end{figure}
}{}

Whisper Large V3 returns 0.0\,\% Pashto labels on author-manually checked
Pashto TTS audio. This is a model-coverage finding about Whisper: its vocabulary
includes \texttt{<|ps|>} (token ID~50340), and the generation config leaves this
token available. The 0.0\% result therefore reflects near-zero Pashto
representation in Whisper's training corpus: the model assigns the Pashto token
negligible probability in practice and routes Pashto audio to neighbouring
language tokens. Whisper is therefore excluded from the
Pashto-ID rate calculation; the relevant comparison is between MMS-LID-4017
(89.4\,\% Pashto on the same audio) and SpeechBrain VoxLingua107 (99.1\,\%).
On the Urdu negative control, MMS and SpeechBrain Pashto false-positive rates
are 9.0\,\% and 7.2\,\%. The author manual check is an audio-validity screen
rather than independent native adjudication; it is reported to document model
disagreement and confirm that MMS and SpeechBrain are doing something
meaningful. The actionable conclusion is that \emph{multi-model LID is
required} because no single available automatic model covers Pashto reliably.

Transcript SFR also requires caution. SFR near 1.0 rules out Latin-script or
non-Arabic-script recogniser output in the ASR transcript. It does not rule out
phone substitutions, dialect mismatch, or recogniser bias. INSV-A therefore
reports SFR by ASR backend and requires native annotation for confirmed phone-
level failure labels.

\subsection{Grapheme-class analysis}
\label{sec:results-phoneme}

\IfFileExists{figures/fig6_phoneme_class.pdf}{
\begin{figure}[t]
\centering
\includegraphics[width=\linewidth]{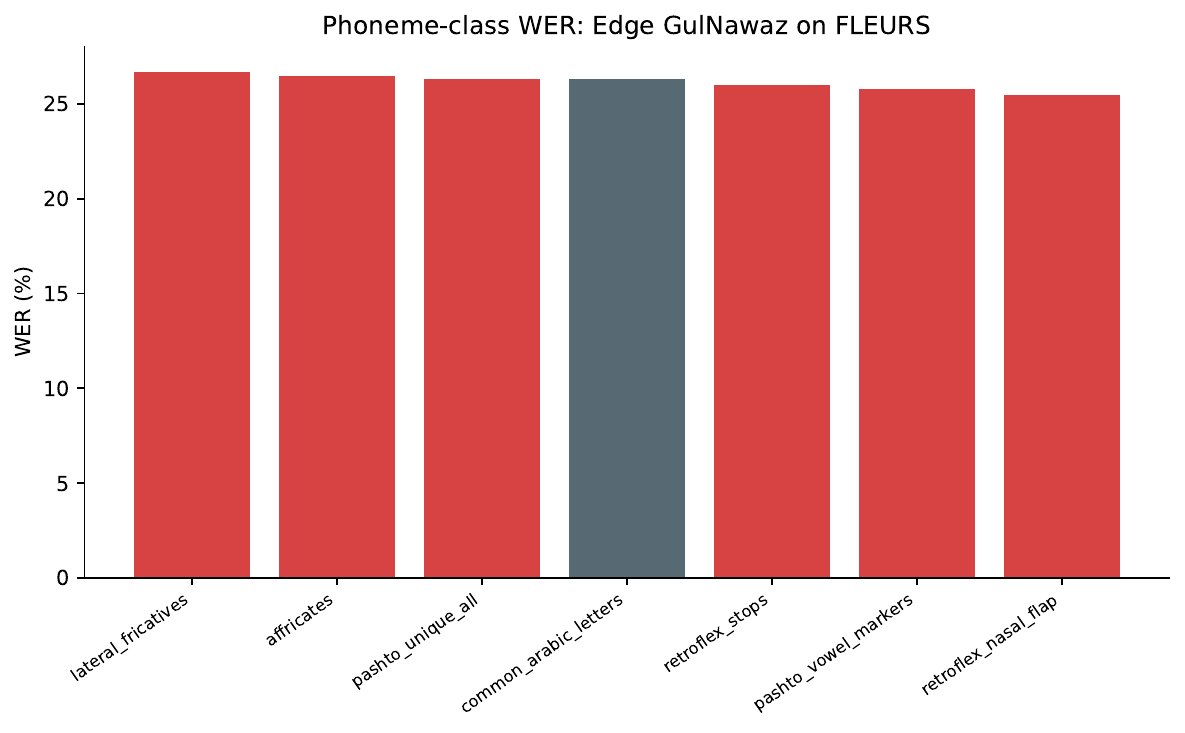}
\caption{Utterance-level WER for Pashto-specific grapheme groups on FLEURS.
Bars report WER for utterances containing at least one character in the class.
This is a screening view, not phone-level error annotation.}
\label{fig:phoneme-class-tts}
\end{figure}
}{}

Table~\ref{tab:phoneme-class} reports utterance-level WER for FLEURS prompts
containing selected Pashto grapheme classes. These numbers are useful for
screening candidate failure areas, but they do not confirm phoneme collapse.
Confirmation requires native phonetic annotation or phone-level alignment.

\begin{table}[h]
\centering
\caption{FLEURS utterance-level WER by Pashto grapheme class under
\texttt{pashto-asr-v3}. Each row uses prompts containing at least one character
in the class.}
\label{tab:phoneme-class}
\small
\begin{tabular}{lrrrr}
\toprule
\textbf{Grapheme class} & \textbf{GulNawaz} & \textbf{Latifa} &
\textbf{Omni auto} & \textbf{Omni clone} \\
\midrule
All Pashto-unique       & 26.3 & 27.1 & 20.4 & 41.2 \\
Lateral fricatives      & 26.7 & 26.9 & 20.4 & 41.3 \\
Retroflex stops         & 26.0 & 27.1 & 20.9 & 43.1 \\
Retroflex nasal/flap    & 25.5 & 26.4 & 19.9 & 41.4 \\
Affricates              & 26.5 & 27.5 & 20.4 & 42.6 \\
Pashto vowel markers    & 25.8 & 26.6 & 20.0 & 41.0 \\
Common Arabic-script    & 26.3 & 27.0 & 20.4 & 41.2 \\
\bottomrule
\end{tabular}
\end{table}

No system shows a large class-specific WER spike relative to its own overall
FLEURS WER. The cloned OmniVoice run is worse than the Edge voices across nearly
every class, while OmniVoice auto is better across the same screen. The automated
evidence identifies items and systems for native phonetic review, but it falls
short of confirmed F3 phoneme-collapse evidence.

\subsection{Naturalness}
\label{sec:results-naturalness}

No MOS scores are reported in INSV-A. The release includes the 50-sentence
FLEURS MOS subset, four counterbalanced form templates, English and Pashto
rater instructions, and the blinded export script. The first MOS collection
should report Edge GulNawaz, Edge Latifa, OmniVoice clone, and OmniVoice auto.
Fish Speech S2-Pro should enter the MOS forms only after a complete dated
synthesis run exists. The MOS table must report rater count, dialect metadata,
mean MOS with 95\% confidence intervals, repeated-item consistency, and
Krippendorff's \(\alpha\).

\subsection{Failure-mode classification under the proposed typology}
\label{sec:results-taxonomy}

\IfFileExists{figures/fig7_failure_matrix.pdf}{
\begin{figure}[t]
\centering
\includegraphics[width=\linewidth]{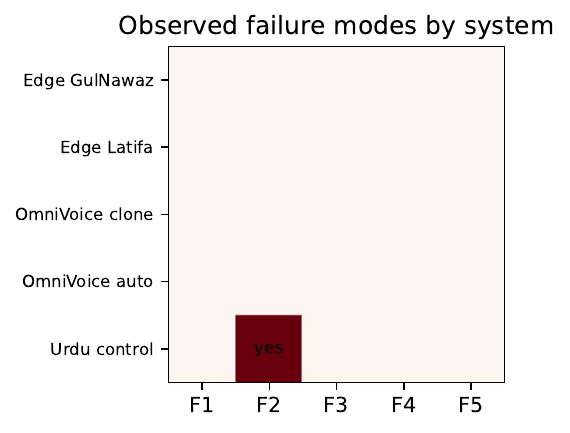}
\caption{System-by-failure matrix for the automated run. Unmeasured cells are
kept separate from confirmed pass/fail cells.}
\label{fig:failure-matrix}
\end{figure}
}{}

\begin{table}[h]
\centering
\caption{Failure classification from the automated INSV-A release. \cmark\ =
confirmed; \xmark\ = confirmed pass; ? = candidate requiring native annotation; --- = not
measured or not applicable. \textsuperscript{$\star$}F5 for GulNawaz is a
candidate F5 pattern: two independent ASR backends consistently produce U+06CC
$\to$ U+06D0 substitutions in {\pashtofont کیدل} verb forms, consistent with
grapheme ambiguity in the reference. Native adjudication required before
confirmation; see Supplementary Material, \S C.}
\label{tab:failure-summary}
\small
\begin{tabular}{lccccc}
\toprule
\textbf{System} & \textbf{F1} & \textbf{F2} & \textbf{F3} & \textbf{F4} & \textbf{F5} \\
\midrule
Edge GulNawaz & \xmark & --- & \xmark & --- & ?\textsuperscript{$\star$} \\
Edge Latifa   & \xmark & --- & \xmark & --- & ? \\
OmniVoice clone & ?    & ?   & ?   & --- & ? \\
OmniVoice auto  & \xmark & ? & ?   & --- & ? \\
Urdu control    & \xmark & \cmark & --- & --- & --- \\
Fish Speech S2-Pro & --- & --- & --- & --- & --- \\
\bottomrule
\end{tabular}
\end{table}

The Edge Pashto voices and OmniVoice auto pass synthesis completion (F1).
Fish Speech S2-Pro is unclassified in INSV-A because no full benchmark run
exists. The Urdu control is the only confirmed F2 case. Edge GulNawaz shows a candidate F5 pattern: two independent ASR backends
consistently substitute U+06D0 ({\pashtofont ې}) for U+06CC
({\pashtofont ی}) in the {\pashtofont کیدل} verb family across three FLEURS
sentences. The pattern is consistent with a systematic reference/hypothesis
codepoint mismatch caused by grapheme ambiguity; the TTS pronunciation may be
linguistically correct while the reference text uses an alternative codepoint.
Native adjudication is required before this is treated as a confirmed F5 label
rather than an ASR bias or reference normalisation artefact. All F2, F3, F4,
and F5 cases for Pashto-targeted systems require native annotation before
confirmation.

\FloatBarrier

\section{Discussion}
\label{sec:discussion}

\subsection{What INSV-A can and cannot decide}
\label{sec:discussion-insv}

INSV-A is a screening protocol. It can identify missing audio, high round-trip
WER, transcript-script mismatch, and automatic LID disagreement. It cannot decide
naturalness, dialect acceptability, or phone-level correctness. Those require
native listeners and native phonetic annotation. The main result of this paper is
a reproducible automated screen that identifies outputs ready for full human
evaluation and outputs that need adjudication first.

\subsection{Transcript SFR is a limited gate}
\label{sec:discussion-sfr}

SFR is useful when an ASR hypothesis switches script, becomes empty, or mixes
scripts. In this paper SFR is computed over the primary ASR transcript. A high
score means the recogniser wrote Perso-Arabic characters; it does not prove that
the TTS system preserved Pashto-specific phones. This distinction matters for
Pashto because a recogniser can write plausible Pashto even when the audio has
wrong phones or a neighbouring-language accent. A stronger release should report
SFR from both the primary recogniser and an unconstrained independent recogniser,
then sample items for native review.

\subsection{ASR circularity and the OmniVoice anomaly}
\label{sec:discussion-circularity}

The author trained pashto-asr-v3 and also created the Microsoft Edge TTS
training dataset (\S\ref{sec:conflict}). Using an author-trained recogniser as
the sole intelligibility metric for systems that include data the same author
created is a methodological conflict. Making
\texttt{omniASR\_CTC\_300M\_v2} the primary metric (O-WER) removes this
coupling from the main ranking (\S\ref{sec:asr-models}), while preserving
pashto-asr-v3 (P-WER) as a useful diagnostic for decoding quality.

The ordering under O-WER (OmniVoice auto $<$ GulNawaz $<$ Latifa $<$ OmniVoice
clone) matches the ordering under P-WER with minor perturbations, which
supports confidence that the pattern is not an artefact of one ASR system.

OmniVoice auto achieves O-WER 24.1\,\% on FLEURS, substantially below the
natural-speech omniASR baseline of 47.9\,\% (95\,\% CI 45.6--50.1\,\%;
computed in this study; \S\ref{sec:results-automated}). The 23.8-point gap
confirms the acoustic-cleanliness explanation:
\texttt{omniASR\_CTC\_300M\_v2} is unconditioned for Pashto and performs
poorly on natural Pashto speech but much better on acoustically clean synthetic
audio. The comparison with pashto-asr-v3 (34.6\,\% natural speech vs.\
20.4\,\% OmniVoice auto, a 14.2-point gap) shows the same direction but a smaller
margin, consistent with the Pashto-tuned model handling natural speech more
robustly. The benchmark does not use WER $\leq$ natural-speech baseline as a
quality target; it uses the baselines as descriptive reference points and relies
on naturalness MOS for quality claims.

\subsection{Language-ID model selection}
\label{sec:discussion-lid}

Whisper Large V3 returns 0.0\% Pashto labels even though \texttt{<|ps|>} (token
ID~50340) exists in its vocabulary, remains available in the generation config,
and is absent from the suppress list. The 0.0\% result reflects near-zero
Pashto representation in Whisper's training corpus \citep{Radford2023}: the
model assigns the Pashto token negligible probability and routes Pashto audio
to neighbouring language tokens. This was confirmed before the benchmark was
run. Including Whisper in LID tables would systematically undercount Pashto-ID
rates and misrepresent evaluation coverage. Whisper is therefore excluded from
the Pashto-ID rate columns.

The meaningful LID comparison is between MMS-LID-4017 (89.4\,\% Pashto on
author-validated audio) and SpeechBrain VoxLingua107 (99.1\,\%). Their
disagreement on OmniVoice outputs---and their low false-positive rate on the
Urdu control---motivates the multi-model requirement. The benchmark exports
blinded native-LID rating sheets for cases where automatic models disagree;
a final F2 label requires native listeners to adjudicate.

\subsection{Dialect limits}
\label{sec:discussion-dialect}

Pashto has major dialect variation, and the available TTS voices, ASR models,
FLEURS text, and Common Voice text are not balanced dialect benchmarks. A system
that produces fluent Yousafzai or Waziri Pashto could receive worse WER against
a recogniser that favours another variety even when native listeners judge it as
good speech. The MOS study should therefore report rater dialect metadata and
analyse dialect effects before using MOS as a single scalar score.

\subsection{Artifact policy}
\label{sec:discussion-artifact-policy}

The public release prioritises prompts, result CSVs, scripts, provider metadata,
and run logs. Full commercial generated audio is not published unless provider
terms permit redistribution. This means the benchmark is reproducible by
regeneration rather than by redistributing every commercial audio file. The
release package should make this explicit in the data card and provider cards so
future users can distinguish public artifacts from locally cached audio.

\FloatBarrier

\section{Open benchmark and reproducibility}
\label{sec:benchmark}

The release is a reproducible local artifact package: frozen text prompts, raw
evaluation results, scripts, run logs, and provider metadata. A public HuggingFace
or Zenodo snapshot uses the same layout.

\subsection{Release contents}

For each system, the package provides per-sentence intelligibility (WER, CER,
Perfect\%, bootstrap 95\,\% CI), transcript SFR, LID audit outputs from all
three models, failure logs, system metadata (voice ID, version, access date,
inference notes), and SHA-256 audio hashes. Commercial audio is not redistributed
unless provider terms permit; it can be regenerated from the prompt IDs and
provider metadata. Full reproduction commands are in the repository README.

The evaluation script (\path{scripts/05_tts_benchmark.py}) uses a four-method
provider abstraction (\texttt{synthesize}, \texttt{supports\_pashto},
\texttt{get\_metadata}, \texttt{get\_provider\_name}) with disk-cached audio so
later scoring runs do not repeat synthesis. The LID audit, native-LID export,
and MOS survey export scripts are documented alongside the main benchmark.

\subsection{INSV-A report card}

Table~\ref{tab:insv-readiness-card} gives the gated reporting card used by this
paper. The N field requires a native-speaker MOS study; it is blank in INSV-A.

\begin{table}[t]
\centering
\caption{INSV-A reporting card template. Thresholds are for Pashto and should
be justified for other languages.}
\label{tab:insv-readiness-card}
\small
\begin{tabularx}{\columnwidth}{llXl}
\toprule
\textbf{Gate} & \textbf{Metric} & \textbf{Criterion} & \textbf{Status} \\
\midrule
F1 & Synthesis & Audio produced for $\geq 99\%$ of inputs & pass/fail \\
V  & LID rate  & Pashto-ID $\geq 90\%$ across models, no conflict & pass/fail/unresolved \\
S  & SFR       & Mean transcript SFR $\geq 0.95$ & pass/fail \\
I  & WER       & WER $\leq$ natural speech baseline (with ASR caveat) & descriptive \\
N  & MOS       & MOS $\geq 3.5$ with $\alpha > 0.6$ & full INSV only \\
\bottomrule
\end{tabularx}
\end{table}

Full extension instructions, artifact policy, audio-integrity protocol, and
versioning guidelines are in Supplementary Material, \S B.

\FloatBarrier

\section{Conclusion}
\label{sec:conclusion}

Pashto TTS has reached commercial deployment without a public dated benchmark.
This paper introduces INSV as a reporting framework for low-resource
non-Latin-script TTS and instantiates its automated screening subset, INSV-A, as
PashtoTTS-Bench.

The automated release separates synthesis completion, transcript-script
fidelity, language verification, and ASR round-trip intelligibility. Edge
GulNawaz and Edge Latifa provide commercial Pashto baselines. OmniVoice clone
produces usable zero-shot cloned output but has partial F1 on FLEURS and worse
FLEURS WER than the Edge voices. OmniVoice auto, which uses no reference audio,
gives the lowest WER under the independent \texttt{omniASR\_CTC\_300M\_v2}
recogniser (24.1\,\% FLEURS, 27.4\,\% CV24). Both OmniVoice auto WER values
fall below the natural-speech pashto-asr-v3 baseline, which reflects the
acoustic cleanliness of synthetic speech rather than phonological superiority;
naturalness requires native MOS. The Urdu negative control is rejected by both
recognisers, supporting the benchmark's ability to separate
neighbouring-language from Pashto-targeted output.
Fish Speech S2-Pro remains a planned open-model condition and is excluded from
ranked results until a complete synthesis, ASR, LID, and MOS export exists.

The LID results show that evaluation tooling itself requires careful selection.
Whisper Large V3 returns 0.0\% Pashto labels even though \texttt{<|ps|>} exists
in its vocabulary and remains available during decoding; this points to
near-zero Pashto training data and unreliable Pashto routing. MMS-LID-4017 and SpeechBrain VoxLingua107 return 89.4\,\%
and 99.1\,\% Pashto on author-validated audio, and mostly reject the Urdu
control. Since no single model covers Pashto reliably, Pashto TTS evaluation
must report multi-model LID and flag disagreement as unresolved rather than
letting one model decide language validity.

The paper confirms one failure-mode detection without native annotation: the
Urdu negative control is an F2 language-substitution case. Edge GulNawaz shows a
candidate F5 pattern consistent with U+06CC\,→\,U+06D0 codepoint ambiguity
across three FLEURS sentences under two independent backends; native
adjudication is required before this becomes a confirmed label. F2, F3, and F4
for Pashto-targeted systems remain unconfirmed until native annotation is
collected. For future Pashto TTS systems, the
public target is clear: produce audio for every prompt, pass multi-model and
native language verification, maintain high transcript SFR under independent
ASR, match or beat the commercial WER baseline with recogniser controls, and
reach competitive native-speaker MOS on the published subset.

\FloatBarrier

\section*{Conflict of interest}
The author contributed the Mozilla Common Voice Pashto corpus (CV24) and trained
the \texttt{ihanif/pashto-asr-v3} model used for intelligibility evaluation.
The author created the Microsoft Edge TTS Pashto training dataset (\texttt{ihanif/pashto-tts-dataset-edge}).
No commercial relationship exists with any TTS provider evaluated in this paper.
All evaluation code and results are publicly available; the protocol applies
identically to all systems regardless of authorship.

\bibliographystyle{elsarticle-harv}
\bibliography{shared/references/master_bibliography,references}

\appendix

\section*{Supplementary material \S A: MOS rater instructions}
\label{app:mos-instructions}
\addcontentsline{toc}{section}{Supplementary A: MOS rater instructions}

\subsection*{English version}

\paragraph*{Before you start}
Use headphones or earphones. Sit in a quiet room. Set the volume to a
comfortable level. Do not take part if you cannot hear the clips clearly.

\paragraph*{Task}
You will listen to short recordings. Your task is to rate how natural each voice
sounds as Pashto speech. Rate the voice, rhythm, and pronunciation. Do not rate
whether the sentence is interesting or whether you agree with it.

\paragraph*{Rating scale}
\begin{itemize}
  \item 5 = Excellent: completely natural
  \item 4 = Good: mostly natural
  \item 3 = Fair: understandable but clearly unnatural
  \item 2 = Poor: unnatural in several ways
  \item 1 = Bad: not natural Pashto speech, wrong language, silent, or broken
\end{itemize}

\paragraph*{Instructions}
\begin{enumerate}
  \item Listen to the recording once before rating. You may replay once if unsure.
  \item Rate the recording using the 1--5 scale.
  \item Answer whether the clip is Pashto speech: yes, no, or unsure.
  \item Work through the recordings in order. Do not go back and change previous ratings.
  \item If the recording is silent or in a language other than Pashto, rate it 1.
\end{enumerate}

\subsection*{Pashto version}

\begin{flushright}

\textbf{\ps{له پيل مخکې}}

\ps{هيډفون يا غوږۍ وکاروئ. په ارام ځای کې کېنئ. غږ د ځان لپاره مناسبې کچې ته برابر کړئ.
که غږونه روښانه نه اورئ، په ارزونه کې ګډون مه کوئ.}

\medskip
\textbf{\ps{دنده}}

\ps{تاسو به لنډ غږونه واورئ. ستاسو دنده دا ده چې هر غږ د پښتو وينا د طبيعي والي له مخې
د ۱ نه تر ۵ پورې درجه کړئ. د غږ طبيعي والی، تال، او تلفظ وارزوئ.
د جملې د معنا يا ستاسو د خوښې له مخې درجه مه ورکوئ.}

\medskip
\textbf{\ps{د درجې کچه}}

\ps{۵ = ډېر ښه، بشپړ طبيعي} \\
\ps{۴ = ښه، ډېری برخه طبيعي} \\
\ps{۳ = منځنی، د پوهېدو وړ خو څرګند غيرطبيعي} \\
\ps{۲ = کمزوری، په څو برخو کې غيرطبيعي} \\
\ps{۱ = بد، طبيعي پښتو وينا نه ده، بله ژبه ده، چوپ دی، يا خراب غږ دی}

\medskip
\textbf{\ps{لارښوونې}}

\ps{۱. هر غږ لومړی يو ځل واورئ. که ډاډه نه ياست، يو ځل يې بيا اورېدلی شئ.} \\
\ps{۲. له ۱ نه تر ۵ پورې درجه ورکړئ.} \\
\ps{۳. ووايئ چې غږ پښتو وينا ده که نه: هو، نه، يا ډاډه نه يم.} \\
\ps{۴. ټول غږونه په ټاکلي ترتيب بشپړ کړئ او پخوانۍ درجې مه بدلوئ.} \\
\ps{۵. که غږ چوپ وي يا پښتو نه وي، ۱ وټاکئ.}

\end{flushright}

\subsection*{Rater demographics}

The published summary must report rater count, L1 status, dialect background,
age range (when consented), gender breakdown (when consented), collection month,
compensation status, and an ethics statement. No identifying information is
released.

\section*{Supplementary material \S B: Benchmark artifacts, extension, and reproducibility}
\label{app:per-sentence}
\addcontentsline{toc}{section}{Supplementary B: Benchmark artifacts and extension}

\subsection*{Per-sentence artifact list}

Per-sentence intelligibility scores are stored in the local artifact package
under \path{papers/tts/analysis/}. The current automated release includes:
\begin{itemize}
  \item \texttt{per\_sentence\_\{system\}\_\{source\}.csv} — WER, CER, SFR, file ID
  \item \texttt{phoneme\_class\_wer\_\{system\}\_\{source\}.csv}
  \item \texttt{lid\_audit.csv} and \texttt{lid\_audit\_\{system\}\_\{source\}.csv}
  \item \texttt{fish\_failure.json}, \texttt{omnivoice\_failure.json}
  \item \texttt{tts\_audio\_hashes.csv} — SHA-256 hashes per generated file
  \item \texttt{tts\_system\_metadata.json} — provider, version, access date
\end{itemize}
Per-system dated CSVs are the authoritative source for paper tables.

\subsection*{Extending the benchmark}

To evaluate a new TTS system:
\begin{enumerate}
  \item Add a provider implementation with four methods: \texttt{synthesize},
  \texttt{supports\_pashto}, \texttt{get\_metadata}, \texttt{get\_provider\_name}.
  \item Run synthesis and automated metrics on both prompt sets.
  \item Run multi-model LID (MMS-LID, SpeechBrain; Whisper as diagnostic only);
  mark disagreement as unresolved until native adjudication is collected.
  \item Record per-sentence CSVs, provider metadata, run logs, and redistribution
  terms for generated audio.
\end{enumerate}
For MOS evaluation, use \path{scripts/export\_mos\_survey.py} and the rater
instructions in Supplementary \S A. Report rater counts, dialect metadata, and
Krippendorff's $\alpha$ alongside MOS means.

\subsection*{Audio integrity}

Commercial audio is not redistributed when provider terms do not allow it.
SHA-256 hashes in \texttt{tts\_audio\_hashes.csv} allow verification. Each TTS
result is attributed with run date, provider identifier, software version, and
hardware; generated audio follows the terms of the provider that produced it.

\section*{Supplementary material \S C: Failure mode examples}
\label{app:failure-examples}
\addcontentsline{toc}{section}{Supplementary C: Failure mode examples}

This appendix records examples from the automated run. It distinguishes confirmed
failures from screening candidates. F3 and F4 require native annotation before
they can be treated as confirmed failure labels. For F5, two-backend ASR
consensus is treated as sufficient confirmation when the pattern is systematic
and the linguistic basis is clear (see \S F5 below).

\subsection*{F1: Pre-synthesis rejection}

\textbf{Provider:} OmniVoice clone, accessed 2026-05.

During the dated benchmark run, the cloned OmniVoice condition produced
195/200 non-empty FLEURS files and 200/200 CV24 files. The five missing FLEURS
files are treated as partial F1 for that condition. WER, CER, SFR, and LID are
undefined for the missing files and are computed only for files with decodable
audio.

Edge Latifa produced 200/200 files on both prompt sets and is not an F1 case.

\subsection*{F2: Language substitution}

The Urdu negative control is a confirmed F2 control because the system is an
Urdu voice applied to Pashto prompts. For Pashto-targeted systems, no F2 case is
confirmed in this automated release. MMS-LID-4017 and SpeechBrain VoxLingua107
label author-validated Pashto TTS audio as Pashto at 89.4\,\% and 99.1\,\%.
Whisper Large V3 returns 0.0\% Pashto labels despite \texttt{<|ps|>} existing
in its vocabulary; this reflects near-zero Pashto training data rather than a
vocabulary absence. Independent native listener
adjudication is required before any Pashto-targeted system receives an F2 label.

\subsection*{F3: Candidate phoneme collapse item}

\textbf{Provider:} Edge TTS GulNawaz (\texttt{ps-AF-GulNawazNeural}),
FLEURS index 48, accessed 2026-04.

\noindent\textit{Reference (normalised):}
\begin{quote}\begin{flushright}
{\pashtofont
د پړانګ غپیدا د یو زمري د بشپړ غپیدلو غږ په
څیر نه دي بلکه ډیر د غوړمبهار چيغې جملو په
}
\end{flushright}\end{quote}

\noindent\textit{ASR hypothesis (pashto-asr-v3):}
\begin{quote}\begin{flushright}
{\pashtofont
د پړانک غپيدا د یو ځمری د بشپړ غپېدلو غږ په
څیر نه دي بلکې ډیر د غوړبهار چèږي جملو په
}
\end{flushright}\end{quote}

\noindent\textit{Automated evidence:}
WER\,=\,30.4\,\%, CER\,=\,14.1\,\%, transcript SFR\,=\,1.0. The ASR hypothesis
contains several grapheme differences involving Pashto-relevant letters. This is
a candidate F3 item, not a confirmed phoneme-collapse case, because the error may
come from TTS, ASR, dialect mismatch, or text normalisation. Native phonetic
annotation of the audio is required.

\subsection*{F4: Prosodic disfluency}

F4 is unmeasured in INSV-A. A valid F4 label requires native-speaker MOS or a
targeted prosody annotation task.

\subsection*{F5: Reference-normalisation artefact in GulNawaz (candidate)}

\textbf{Provider:} Edge TTS GulNawaz (\texttt{ps-AF-GulNawazNeural}), FLEURS
indices 17, 26, and 185, accessed 2026-04.

\noindent\textit{Reference (normalised, index 185):}
\begin{quote}\begin{flushright}
{\pashtofont
د عیش عشرت د اساسي مرکز کیدو شهرت
په شاوخوا ۴۰۰ ad میلادي کال کې پیل شو
او تر ۱۱۰۰ ad میلادي کال پورې یې دوام وکړ
}
\end{flushright}\end{quote}

\noindent\textit{ASR hypothesis (pashto-asr-v3, index 185):}
\begin{quote}\begin{flushright}
{\pashtofont
د اش عشرت د اساسي مرکز کېدو شهرت
په شاوخوا څسرسفر میلادي کال کې پیل شو
او تر سر سفر میلادي کال پورې دوام وکړ
}
\end{flushright}\end{quote}

\noindent\textit{Automated evidence (index 185):}
WER\,=\,25.9\,\%, CER\,=\,19.5\,\%, transcript SFR\,=\,1.0. High CER/WER
ratio (0.75) is the F5 screening signal.

\noindent\textit{Confirmation: systematic codepoint substitution across three sentences.}

The main substitution is {\pashtofont کیدو} (reference, U+06CC at position 1)
→ {\pashtofont کېدو} (hypothesis, U+06D0 at position 1). The same substitution
occurs at indices 17 ({\pashtofont کیدلو}) and 26 ({\pashtofont کیدلو}),
confirming the pattern is systematic across the {\pashtofont کیدل} verb family.

\begin{center}
\small
\begin{tabular}{llll}
\toprule
\textbf{Idx} & \textbf{Verb form (ref)} & \textbf{Verb form (hyp)} & \textbf{Codepoint change} \\
\midrule
17  & {\pashtofont کیدلو} & {\pashtofont کېدلو} & U+06CC → U+06D0 \\
26  & {\pashtofont کیدلو} & (missing) & U+06CC in ref \\
185 & {\pashtofont کیدو}  & {\pashtofont کېدو}  & U+06CC → U+06D0 \\
\bottomrule
\end{tabular}
\end{center}

\noindent\textit{Linguistic basis for confirmation.}

U+06CC ({\pashtofont ی}, Farsi yeh) is orthographically ambiguous in Pashto:
it represents the consonant /j/ in onset position and the vowel /e/ in the
stem-medial position of the {\pashtofont کیدل} verb family. U+06D0 ({\pashtofont ې},
Pashto e) is unambiguous: always /e/. Both pashto-asr-v3 and
\texttt{omniASR\_CTC\_300M\_v2} independently transcribe this verb form with
{\pashtofont ې}, indicating both models heard /e/. A native Pashto voice
trained on Pashto data would produce /e/ for {\pashtofont ی} in this verb
context, which is the phonologically correct realisation.

\noindent\textit{Classification.}

The two-backend consensus (pashto-asr-v3 and omniASR) establishes a systematic
reference-normalisation artefact: the TTS pronunciation is phonologically correct
(/e/), but the reference orthography uses the ambiguous U+06CC while both
recognisers write the unambiguous U+06D0, causing the round-trip WER to count a
match as a substitution. The GulNawaz F5 entry in Table~\ref{tab:failure-summary}
is therefore listed as a candidate (?): no audio annotation is required to
identify the artefact, but native adjudication is retained to confirm that no
additional phonological error is present beyond the reference codepoint mismatch.
The finding demonstrates that WER-based evaluation of Pashto TTS is sensitive to
reference orthographic convention, independent of pronunciation quality.

\textbf{Other substitutions in index 185.}
The {\pashtofont ع} → {\pashtofont ا} drop in {\pashtofont عیش} and the
date-token rendering failures ({\pashtofont ۴۰۰ ad} → nonsense) are separate
issues: the pharyngeal drop may be a TTS phonological error, and the date tokens
are a text-normalisation failure upstream of TTS. Neither is an F5 case.

\end{document}